\pdfoutput=1

\documentclass[11pt]{article}

\usepackage[]{ACL2023}

\usepackage{times}
\usepackage{latexsym}

\usepackage[T1]{fontenc}

\usepackage[utf8]{inputenc}

\usepackage{microtype}

\usepackage{inconsolata}
\usepackage{multirow}
\usepackage{graphicx}
\usepackage{spverbatim}
\usepackage{amsmath} 
\usepackage{url}
\title{EWEK-QA: \underline{E}nhanced \underline{W}eb and \underline{E}fficient \underline{K}nowledge Graph \underline{R}etrieval for Citation-based \underline{Q}uestion \underline{A}nswering Systems}

\author{Mohammad Dehghan$^1$, {\bf Mohammad Ali Alomrani$^1$,} {\bf Sunyam Bagga$^1$,} \\ {\bf David Alfonso-Hermelo$^1$,} {\bf Khalil Bibi$^1$,} {\bf Abbas Ghaddar$^1$,} {\bf Yingxue Zhang$^1$,} \\ {\bf Xiaoguang Li$^1$,} {\bf Jianye Hao$^1$,} {\bf Qun Liu$^1$,} {\bf Jimmy Lin$^2$,} {\bf Boxing Chen$^1$,} \\ {\bf Prasanna Parthasarathi$^1$,} {\bf Mahdi Biparva$^1$,} {\bf Mehdi Rezagholizadeh$^1$} \\
  $^1$ Huawei Noah's Ark Lab, $^2$ University of Waterloo \\
  \texttt{\{mohammad.dehghan@uwaterloo.ca, mehdi.rezagholizadeh@huawei.com\}} \\}

\newcommand{\ours}{EWEK-QA }
\newcommand{\ourret}{Adaptive retriever }

\begin{document}
\maketitle
\begin{abstract}
The emerging \textit{citation-based QA} systems are gaining more attention especially in generative AI search applications. The importance of extracted knowledge provided to  these systems is vital from both accuracy (completeness of information) and efficiency (extracting the information in a timely manner). 
In this regard, citation-based QA systems are suffering from two shortcomings. First, they usually rely only on web as a source of extracted knowledge 
and adding other external knowledge sources can hamper the efficiency of the system. Second, web-retrieved contents are usually obtained by some simple heuristics such as fixed length or breakpoints which might lead to splitting information into pieces. To mitigate these issues, we propose our enhanced web and efficient knowledge graph (KG) retrieval solution (EWEK-QA) to enrich the content of the extracted knowledge fed to the system. This has been done through designing an adaptive web retriever 
and incorporating KGs triples in an efficient manner. 
We demonstrate the effectiveness of \ours over the open-source state-of-the-art (SoTA) web-based and KG baseline models using a comprehensive set of quantitative and human evaluation experiments. 
Our model is able to: first, improve the web-retriever baseline in terms of extracting more relevant passages (>20\%), the coverage of answer span (>25\%) and self containment (>35\%); second, obtain and integrate KG triples into its pipeline very efficiently (by avoiding any LLM calls) to outperform the web-only and KG-only SoTA baselines significantly in 7 quantitative QA tasks and our human evaluation.

\end{abstract}

\begin{figure}[ht]
    \centering
    \includegraphics[width=\linewidth, trim = 6cm 2cm 6cm 2cm, clip]{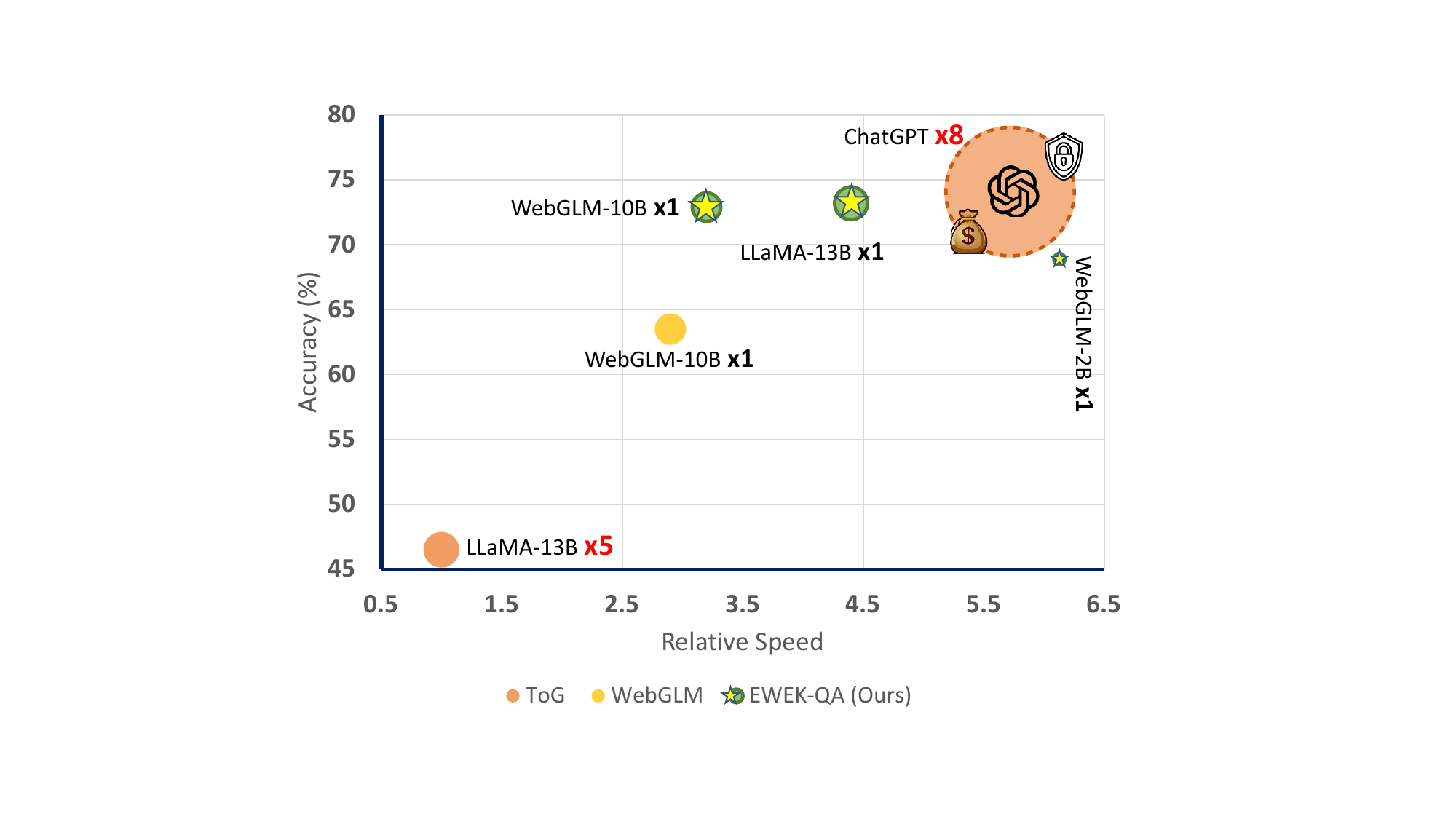}
    \caption{ Overview of performance vs. efficiency of \ours (KG+Web), WebGLM (Web-only), and ToG (KG-only) on the WebQSP dataset (See Table~\ref{tab:efficiency} for details). Each circle represents one solution with its LLM's name and the number of calls to the LLM (indicated as $\times n$). The size of each circle indicates the relative size of its corresponding backbone LLM. The relative speed represents the speed with respect to ToG with {LLaMA-2}-13B. Bear in mind that ToG with ChatGPT needs to call the closed-source ChatGPT system 8 times on average, which can increase the expenses and also raise privacy concerns for sensitive applications.   
    }
    \label{fig:performance_vs_speed}
\end{figure}

\section{Introduction}
Large language models (LLMs) have shown great potentials to be used for question answering (QA)~\cite{chatgptforqa} tasks. However, relying only on the internal knowledge (gained from pre-training or fine-tuning) of LLMs for answering questions may lead to issues such as hallucination, lack of knowledge, or outdated knowledge~\cite{gao2023retrieval}. To address these problems, retrieval augmented generation (RAG)~\cite{gao2024retrievalaugmented} can be leveraged to assist with grounding the answer of LLMs to some external knowledge bases such as web or knowledge graphs (KGs). 
While this approach can be very useful in practice to reduce the hallucination of LLMs~\cite{huang2023a}, it still remains challenging to identify which parts of the answer come from the external knowledge or internal knowledge of LLMs (i.e. knowledge grounding). 

Citation-based QA systems, such as generative AI search applications (e.g. Microsoft's new Bing~\footnote{\url{https://www.microsoft.com/en-us/edge/features/bing-chat?form=MA13FJ}} or YOU.com~\footnote{\url{https://you.com/?chatMode=default}}), aim at addressing the knowledge grounding issue by adding proper citations from relevant retrieved passages (so-called \textit{quotes} in this paper hereafter) to their answer.

In this regard, instruction-tuned LLMs learn to give citations by supervised fine-tuning or in-context learning~\cite{webglm}.

Moreover, considering the abundant number of users and queries to these citation-based QA systems, the whole pipeline should be designed to run very efficiently while providing accurate answers.

A case in point is WebGLM~\cite{webglm}

which is an efficient web-enhanced question answering system based on the 10B GLM model~\cite{du-etal-2022-glm}. 
To the best of our knowledge, WebGLM is the first of its kind to efficiently use open-source models for QA systems with citation capability. In this paper, we aim to improve the accuracy of WebGLM while keeping its efficiency. 
While WebGLM~\cite{webglm} outperforms a similar size (13B) WebGPT model~\cite{nakano2022webgpt} significantly and works on-par with the large WebGPT model (175B), there remains some major challenges to deal with. First, 
WebGLM only relies on the web as a source of external knowledge~\cite{webglm}, which might not be sufficient on its own for answering a diverse set of questions (e.g. multi-hop reasoning questions~\cite{yang2018hotpotqa} or knowledge graph question answering (KGQA)~\cite{perevalov-etal-2022-knowledge} tasks). Second, its web-retrieval module, usually breaks the pages based on some simple heuristics such as length or breakpoints (to form the quotes), which can give rise to splitting complete information into independent pieces.

To address these problems, we propose our \textit{E}nhanced \textit{W}eb and \textit{E}fficient \textit{K}nowledge graph retrieval for citation-based \textit{QA} systems (\ours) which tries to enrich the content of the extracted quotes fed to the LLM in WebGLM through incorporating KGs and extracting adaptive quotes rather than fixed-length quotes from the web. It is worth mentioning that state-of-the-art (SoTA) KGQA techniques~\cite{tog,luo2024rog}, which extract informative triples from KGs, require several calls to LLMs. In this regard, using open-source LLMs can significantly increase the end-to-end latency of the models and using closed-source LLMs (such as ChatGPT) can bring-up extensive costs and privacy concerns (see Fig.~\ref{fig:performance_vs_speed}).
Additionally, the sheer size of modern KGs (e.g. Freebase) makes it challenging to efficiently extract the most relevant sub-graphs. Hence, our solution focuses on retrieving the most informative triples from KGs with \textit{minimal} calls to the LLMs (to maintain the efficiency of the entire pipeline).

We evaluate our \ours using several qualitative (i.e. human evaluations) and quantitative experiments on different types of QA tasks such as open-domain QA (ODQA), multi-hop reasoning, and KGQA. The results show that our adaptive web-retriever

can significantly improve the quality of extracted quotes in terms of relevance to the queries, coverage of the answer span, and self-containment (i.e. containing complete information to answer questions). 
\ours with its efficient graph retriever and adaptive web-retriever is able to outperform both WebGLM and SoTA Think-on-Graph (ToG)~\cite{tog} significantly on KGQA and ODQA datasets by >10\% and >3\% on average respectively. Moreover, \ours achieves between $\sim3\times$ to $\sim6\times$ speedup compared to ToG when using open-source {LLaMA-2}-13B model~\cite{touvron2023llama} (see Fig.~\ref{fig:performance_vs_speed}). Finally, our human evaluation shows that \ours answers questions >20\% more accurately than SoTA baselines.   
The results indicate the importance of combining web-extracted knowledge with KG-extracted triples in designing citation-based QA systems. 
Our contributions are summarized as follows:
\begin{itemize}
    \item We propose an efficient citation-based QA system that utilizes two external knowledge modalities: web text, and KGs simultaneously without hampering the efficiency. Our efficient KG extraction module does not use any open-source or close-source LLM calls but still the triplets provide valuable information to the system.
    \item \ours introduces an adaptive web-retrieval module which is able to extract more informative and more relevant quotes.

    \item We demonstrate that our solution is able to outperform KG-only and Web-only QA baselines in wider range of QA tasks such as  KGQA, ODQA, and multi-hop reasoning datasets based on comprehensive quantitative and human evaluations. 
\end{itemize}

\begin{figure*}[ht]
    \centering
    \includegraphics[width=1\linewidth, trim = 0cm 0cm 0cm 0cm, clip]{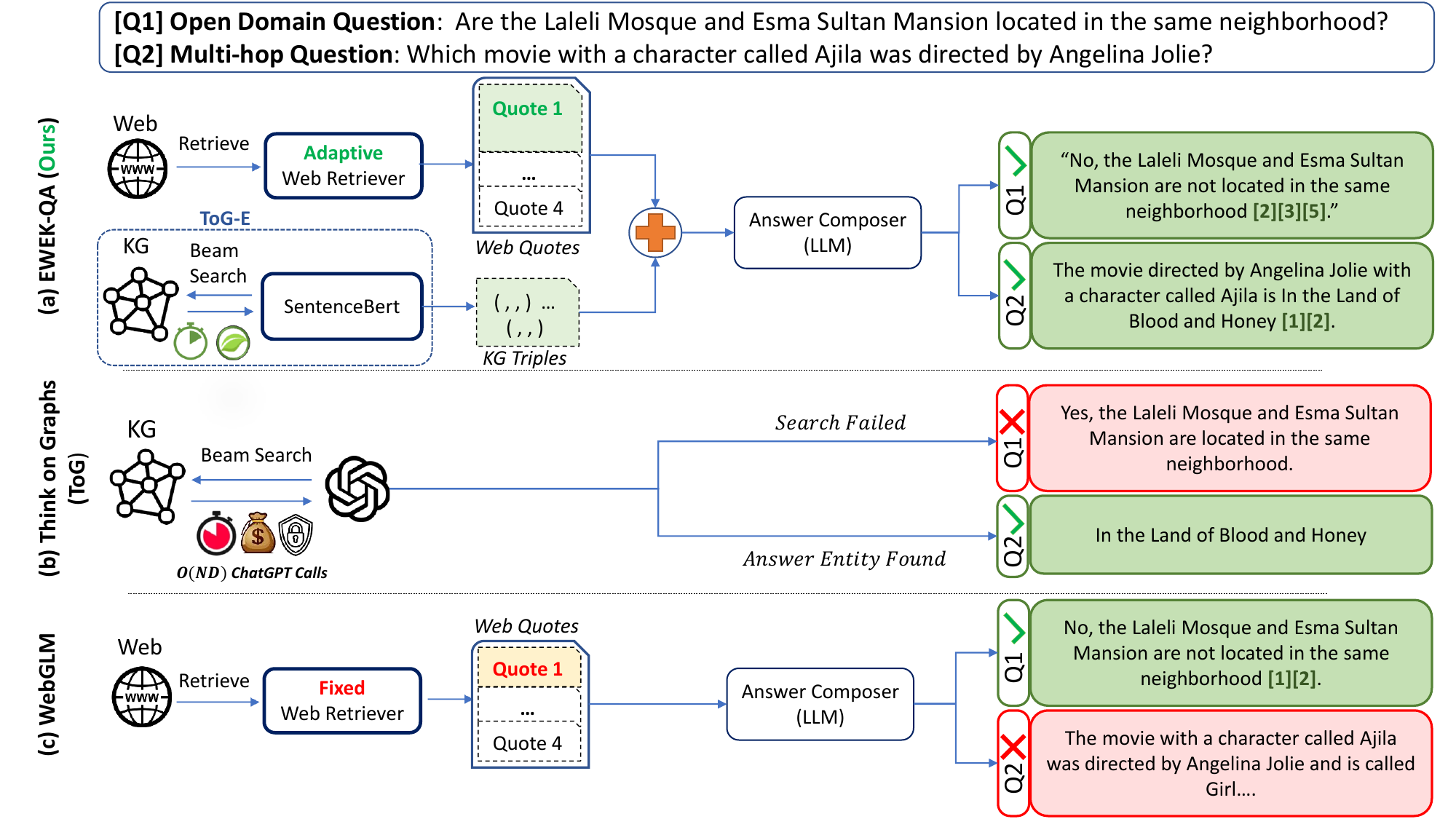}
    \caption{Comparison of \ours, ToG~\cite{tog}, and WebGLM~\cite{webglm} pipelines. \ours utilizes both knowledge modalities which enables to correctly answer both question types using a single LLM call. ToG requires $O(ND)$ costly calls where $N$ and $D$ represent the beam search width and depth respectively. WebGLM relies solely on the web which makes it unfit for multi-hop reasoning questions.}
    \label{fig:webtog}
\end{figure*}

\section{Related Work}
\paragraph{Citation-based Question Answering Systems} 
Citation-based QA systems can be viewed as an enhanced version of regular retrieval augmented generation (RAG) solutions (such as REALM~\cite{realm}, RAG~\cite{rag2020}, and Atlas~\cite{atlasjlmr}) which are able to add citation from relevant retrieved quotes during the answer generation. 
RAG are able to integrate external knowledge into the generation process; however, they cannot add citation to the answers. 
WebGPT~\cite{nakano2022webgpt} is one of the pioneering works on citation-based QA which fine-tuned GPT-3 (175B) to answer open-domain questions using web by browsing through most relevant pages and adding references to the answers from the relevant pages.   
GopherCite~\cite{menick2022teaching} is another case in point where a 280B  model LLM was fine-tuned using reinforcement learning based on human preference to generate answers with proper citations.   
Although the models such as WebGPT and GopherCite rely on the power of large scale LLMs, WebGLM~\cite{webglm} introduced an efficient citation-based QA approach based on fine-tuning much smaller LLMs. WebGLM showed better performance compared to a similar size (13B) WebGPT model~\cite{nakano2022webgpt} significantly and works on-par with the large WebGPT model (175B).
To the best of our knowledge, WebGLM is the first of its kind to efficiently use open-source models for QA systems with citation capability. We found WebGLM as the most relevant work to ours and we keep that as one of our main SoTA baselines. We aim at improving the accuracy of WebGLM while keeping its efficiency.

\paragraph{Knowledge Graph-Augmentation for Reasoning in LLMs }

Many recent works leverage the rich and structured knowledge of KGs to mitigate the hallucination and reliability issues of LLMs. Early studies~\cite{yasunaga-etal-2021-qa, zhang2021greaselm, yasunaga2022dragon} integrate KG embedding methods using GNNs with LLM models during the finetuning or pre-training stage. While such approaches have shown promising results in explainability and reasoning, they tend to require extensive task-specific finetuning.

Another line of works combines external knowledge from KGs into LLMs during the prompting stage. KB-Binder~\cite{li-etal-2023-shot} leverages the Codex LLM~\cite{codex} to generate a SPARQL query that extracts the answer from the KG. KAPING~\cite{Baek2023} retrieves the most relevant one-hop KG triples via dense retrieval models and provides them to the LLM as context. KD-CoT~\cite{wang2023knowledgedriven} proposes to verify and modify CoT reasoning traces with KG knowledge in order to alleviate hallucination and error propagation. MindMap~\cite{wen2023mindmap} builds a prompting pipeline where the LLM reasons over the extracted KG subgraphs and generates answers grounded by the "reasoning pathways" within the KG. 
RoG~\cite{luo2024rog} uses an LLM to retrieve reasoning paths from the KGs based on relation "plans" grounded by KGs.
ToG~\cite{tog} performs beam search on KGs using LLMs to dynamically extract the most relevant reasoning paths. While it displays impressive performance on KGQA datasets, it requires many LLM calls per questions, and it degrades on open-domain datasets where the answer may not exist in the KG. Additionally, it heavily relies on closed-source LLMs (e.g. ChatGPT) for SoTA performance.

\section{Methodology}
\label{sec:3}

Our approach consists of two main components: Knowledge Extraction (\S\ref{sec:3.1}) and Answer Composition (\S\ref{sec:answer_composer}), which operate sequentially.
When presented with a question, we initiate the knowledge extraction phase to gather the pertinent information required for answering. Subsequently, we employ an open-source LLM to generate a cohesive final answer \textit{solely} based on the collected information. See Fig. \ref{fig:webtog} for the full pipeline.

\subsection{Knowledge Extraction: Web and KG}

We concurrently extract information from two distinct knowledge sources -- Web (\S\ref{sec:adaptive_ret}) and Knowledge Graph (\S\ref{sec:tog-e}) -- to gather external data for addressing a given query.
\label{sec:3.1}

\subsubsection{Adaptive Web Retrieval}
\label{sec:adaptive_ret}

We introduce an adaptive module designed to extract pertinent passages (referred to as quotes) from unstructured web text (refer to Figure \ref{fig:adaptive_web_retriever} in the Appendix). Initially, relevant webpages are retrieved using the Bing search engine, followed by a multi-step process to extract quotes, as detailed below. Our approach is termed "adaptive" because it integrates a heuristic-based parser, the Paragraph Splitter (PS), with a small language model, the Evidence Extractor (EE). This dynamic extraction process adjusts according to both the specific query and the format of the webpage, managed by PS and EE respectively. Once webpages are transformed into candidate quotes, our Retrieve and Rerank module selects the most relevant quotes, followed by removal of redundant quotes by a deduplicator module. We provide a detailed explanation of each module below.

\paragraph{Paragraph Splitter (PS).} This module is similar to the WebGLM retriever. As in \citet{webglm}, we divide the webpage contents into a list of candidate passages using line breaks. We apply additional constraints to further improve the quality of the quotes: we utilise \texttt{<p>} tags in the webpage's HTML to produce candidate passages; any passage with less than $10$ tokens is discarded; any passage with more than $80$ tokens is broken down into shorter passages while respecting sentence boundaries (see \S\ref{sec:adaptive-web-appendix} for more details).

\paragraph{Evidence Extractor (EE).} The task is similar to the machine reading comprehension (MRC)~\citep{bert} task. Instead of pursuing an answer span (like in MRC), the target here is to extract \textit{evidence spans} from the webpage contents that can provide support to answer the question. We fine-tune a pre-trained MRC model -- DeBERTa \cite{he2021deberta} -- on the MS Marco dataset \citep{bajaj2016ms} to identify text spans from webpages that are relevant to the user query (see \S\ref{sec:apendix_EE} for more details).

\paragraph{Retrieve and Rerank.} Given all the candidate quotes produced by PS and EE, we retrieve and re-rank them based on semantic relevance using two cross-encoder models of different sizes: (1) a small filtration model to remove irrelevant quotes and (2) a larger cross-encoder model to re-rank the filtered passages. Specifically, we use a six-layer MiniLM \cite{minilm} with $22M$ parameters as the filtration model and a large DeBERTa \cite{he2021deberta} with $900M$ parameters as the re-ranker model. Both models are trained on the MS Marco dataset for the passage ranking task.

\paragraph{Deduplicator.} Since both PS and EE extract quotes from the same webpages, there is a need to eliminate duplicate quotes. We use a small bi-encoder six-layer MiniLM to produce sentence embeddings for each passage and compute the cosine similarity between each pair of embeddings. Any passage with cosine similarity > $0.9$ is removed.

\subsubsection{ToG-E: Sub-graph Retrieval}

\label{sec:tog-e}
Knowledge Graphs (KGs) serve as structured and dynamic repositories of information. Integrating LLMs with KGs presents an adjunctive strategy to mitigate LLM's limitations in answering multi-hop reasoning questions~\cite{tog}. 
Prior KG sub-graph extraction modules demonstrated enhanced effectiveness across diverse question types~\cite{wang2023knowledgedriven, luo2024rog}.

Adhering to the Think-on-Graph (ToG) methodology \citep{tog}, we employ beam search~\citep{jurafskyspeech} on the KG to extract a relevant sub-graph given a question. In a nutshell, ToG iteratively invokes an LLM to explore possible reasoning paths on
the KG until the LLM determines that the question can be answered based on the current reasoning
paths. At each iteration, ToG constantly updates the top-$N$ reasoning paths until a max depth $D$ is reached. To enhance this execution, we introduce an efficient variant of the ToG method, denoted as ToG-E.  In contrast to the original ToG methodology, the ToG-E returns a sub-graph in the form of "entity, relation, entity" triples without invoking any LLM. 

Three primary distinctions characterize ToG-E in comparison to ToG: $1$) During the pruning step, ToG relies on an LLM to acquire scores for candidate relations and entities in the beam search, whereas ToG-E utilizes SentenceBert~\cite{reimers-gurevych-2019-sentence} embeddings of the question, relations, and entities to compute cosine similarity scores for each relation and entity. $2$) ToG-E omits a reasoning step employed by ToG to halt before reaching the maximum depth in the beam search. 3) The ToG-E method produces a sub-graph as its output, regardless of whether the extracted sub-graph is informative or not. In contrast, the ToG methodology validates the extracted sub-graph (triples) by engaging an LLM and may further prompt the LLM with a CoT~\citep{cot} instruction. This additional step aims to elicit an answer solely based on the parametric knowledge of the LLM in case the sub-graph is deemed to lack sufficient informativeness. 

In contrast to ToG, which relies heavily on closed-source LLMs such as ChatGPT to achieve effectiveness, our approach sidesteps the use of any LLM during the triple extraction process from the KG (see Figure \ref{fig:webtog}). Furthermore, our system does not require any KGQA supervised dataset for training or fine-tuning. As a result, we compare our method with prompt-based approaches.

\subsection{Answer Composition}\label{sec:answer_composer}
After extracting KG triples and web quotes, we utilize an open-source pre-trained LLM to process this data and construct a coherent response, supplemented with citations to relevant knowledge sources. 
The KG triples constitute the initial passage, while the subsequent passages consist of web quotes, all serving as inputs to the answer composer model (see \S\ref{app:answer_composer} for details). 

In our experiments, we employed the WebGLM-10B model \citep{webglm} for answer composition by default. This model has been fine-tuned specifically for the task of composing answers: given a question and a set of text passages (5 to 10) containing relevant information, the LLM is trained to produce an accurate answer grounded in these passages.

\begin{table*}[htb!]
\centering
\resizebox{\textwidth}{!}{\begin{tabular}{lccccccc|c}
\hline
\textbf{Method}                                & \textbf{WebQSP} & \textbf{CWQ} & \textbf{WebQuestions} & \textbf{{HotpotQA}} & \textbf{GrailQA} & \textbf{SimpleQA} & \textbf{Natural Questions} & \textbf{Avg.} \\ \hline
\multicolumn{9}{c}{\textit{w/o External Knowledge}}   \\ \hline
\texttt{IO Prompt w/ChatGPT}                   & $59.8$                           & $39.4$                        & $53.7$                                 & $31.2$                           & $27.4$                            & $18.5$                             & $51.1$                                      & $40.1$                         \\ 
\texttt{CoT w/ChatGPT}                         & $61.0$                           & $37.8$                        & $54.1$                                 & $33.1$                           & $29.6$                            & $18.8$                             & $52.8$     & $41.0$          \\ \hline
\multicolumn{9}{c}{\textit{fine-tuned w/External Knowledge}}                               \\ \hline
\texttt{DeCAF}~\cite{yu2023decaf}                                 & $76.6$                           & $56.6$                        & -                                      & -                                & -                                 & -                                  & -                                           & -                              \\
\texttt{KD-CoT}~\cite{wang2023knowledgedriven}                                & $73.7$                           & $50.5$                        & -                                      & -                                & -                                 & -                                  & -                                           & -                              \\ \hline
\multicolumn{9}{c}{\textit{Prompting w/External Knowledge}}                               \\ \hline
\texttt{ToG (ChatGPT)}~\cite{tog}                                   & $74.1$                           & $46.7$                        & $59.3$                                 & $28.3$                           & $70.8$                            & $56.7$                             & $44.8$                                      & $54.3$                         \\
\texttt{WebGLM}~\cite{webglm}                                   & $63.5$                           & $42.3$                        & $54.3$                                 & $38.7$                           &              $34.3$                  &      $29.7$                     &                 $57.6$       &    $45.8$                                    \\
\texttt{\ours} (Ours) w/ KG          & $59.9$                           & $40.1$                        & $50.7$                                 & $20.6$                           & $70.0$                            & $57.9$                             & $17.8$                                      & $45.2$                         \\  
\texttt{\ours} (Ours) w/ Web         & $68.0$                           & $48.1$                        & $58.1$                                 & $42.9$                           & $36.7$                            & $33.0$                             & $64.7$                                      & $50.2$                         \\  
\texttt{\ours} (Ours) w/ KG + Web & $71.3$ (+$\mathbf{7.8}$)         & $52.5$ (+$\mathbf{10.2}$)      & $61.2$ (+$\mathbf{6.9}$)               & $43.6$ (+$\mathbf{4.9}$)         & $60.4$ (+$\mathbf{26.1}$)         & $50.9$ (+$\mathbf{21.2}$)          & $62.5$ (+$\mathbf{4.9}$)                             & $57.4$       \\
\hline
\end{tabular}}
\caption{Hits@1 accuracy $\uparrow$ for different datasets. \textbf{KGQA Datasets}: WebQSP, CWQ, GrailQA, and SimpleQA. \textbf{Open-domain Datasets}: WebQuestions, Hotpot, and Natural Questions. For \ours, we compare having access to only KG triples, only web quotes, or both. The parentheses represent improvement over WebGLM.}
\label{tab:main_table}
\end{table*}

\section{Experiments}

\subsection{Experimental Setup}
\label{sec:experimental_setup}

\paragraph{Datasets.}
We use $4$ KGQA datasets to evaluate the multi-hop reasoning abilities of our approach: WebQSP \cite{yih-etal-2016-value}, CWQ \citep{talmor-berant-2018-web}, GrailQA \cite{grailqa}, and SimpleQA \cite{simpleqa}. Additionally, we evaluate on $3$ ODQA datasets: HotpotQA \cite{yang-etal-2018-hotpotqa}, WebQuestions \cite{berant-etal-2013-semantic}, and Natural Questions (NQ)~\cite{kwiatkowski-etal-2019-natural}. We evaluate our models on a random sample of $1000$ instances from each dataset to manage computational costs. However, for WebQSP and CWQ, we use the full test set, and for Natural Questions, we assess a random subset of 400 samples.

Freebase KG \cite{freebase} is utilized for all datasets. See \S\ref{app:imp_details} for details.

\paragraph{Evaluation Metrics.}

We compute Hits@$1$ to evaluate the models' answers following prior works \citep{Baek2023, Jiang-StructGPT-2022, li-etal-2023-shot}. That is, each question receives a score of $1$ if the target answer is present within the predicted LLM answer, and $0$ otherwise. The metrics used in our human evaluation studies are discussed in \S\ref{sec:final-answer-human-eval} and \S\ref{sec:web-retriever-human-eval}.

\paragraph{Baseline Methods.}

We use standard prompting (IO Prompt) \cite{gpt3} and Chain of Thought (CoT) prompting \cite{cot} with $6$ in-context examples as prompting baselines \textit{with no external knowledge}. Additionally, we compare our method to $4$ SoTA baselines that can access external knowledge: ToG \cite{tog}, KG-CoT~\cite{wang2023knowledgedriven}, DeCaF~\cite{yu2023decaf}, and WebGLM~\cite{webglm}.

ToG performs beam search on KGs using ChatGPT to keep track of the most relevant reasoning paths. KD-CoT generates faithful reasoning traces based on retrieved external knowledge to produce precise answers. DeCaF is a finetuning-based method that jointly generates search queries over KGs and predicts the final answer. Finally, WebGLM is a web-based QA system that composes an answer based on external knowledge retrieved through a web search. 

We experiment with two backbone language models for ToG: ChatGPT (Table \ref{tab:main_table}) and Llama-$2$-$13$B (Table \ref{tab:efficiency}) \cite{touvron2023llama}. For \ours and WebGLM, we use the WebGLM-$10$B answer composer in all experiments unless specified otherwise. 
Refer to \S\ref{app:imp_details} for more details.

\paragraph{Human Evaluation Setup.} 
We perform two human annotation experiments to evaluate the quality of: (i) final answer generated by the model (\S\ref{sec:final-answer-human-eval}) and (ii) web quotes extracted by our adaptive retriever (\S\ref{sec:web-retriever-human-eval}). Four professional annotators are carefully selected from a pool of $20$ candidates based on their demonstrable skills and expertise. They are trained and continuously monitored by our domain expert. We present them with detailed task-specific instructions and illustrative examples covering potential scenarios that they might encounter during the annotation process. They contribute $600$ hours of total annotation time on both tasks and receive a compensation of \$$19$ USD per hour.

\subsection{Main Results}

In this section, we present the end-to-end evaluation of our proposed system. We evaluate the quality of the final generated answer using Hits@1 accuracy (\S\ref{sec:final-answer-automatic-eval}) and human annotators (\S\ref{sec:final-answer-human-eval}).

\subsubsection{Automatic Evaluation}\label{sec:final-answer-automatic-eval}
The performance comparison of \ours with different prompting and finetuning baselines is presented in Table \ref{tab:main_table}. We observe that standard IO prompting results in particularly low performance for multi-hop open-domain and KGQA datasets (e.g. HotpotQA and GrailQA). Interestingly, CoT delivers only a minor improvement over standard prompting which shows that such multi-hop domains require more than relying on the internal parametric knowledge of ChatGPT. Although DeCAF outperforms KG-CoT and exhibits top performance for WebQSP and CWQ, it requires extensive finetuning for each dataset making it an impractical choice for a generic QA system. ToG shows competitive performance on KGQA datasets but falls short on open-domain datasets such as HotpotQA. Note that, unlike the open-domain setting, KGQA methods assume that the answer is an entity that exists in the KG; making them unsuitable for generic questions (e.g. Yes/No questions). 

We observe that \ours with only web quotes surprisingly performs well on some KGQA datasets (e.g. CWQ) and significantly outperforms all baselines on HotpotQA. This demonstrates the power of web-based QA systems for multi-hop open-domain questions which require reasoning over more than one supporting passage to answer. 
Having access to both external knowledge modalities inherits both the benefits of WebGLM and ToG, achieving competitive performance on 5/7 datasets while using a fraction of computational cost (efficiency details in Table \ref{tab:efficiency} and \S\ref{sec:eff_analysis}).

\subsubsection{Human Evaluation} \label{sec:final-answer-human-eval}

\paragraph{Metrics and Data.} 
 
We adopt a $3$-level \textit{Correctness} score inspired by {\citet{gao2023enabling}}. Each model output receives one of three labels: \texttt{IDK} (Does not know or is unable to answer), \texttt{Incorrect} (Is fully or partially incorrect), or \texttt{Correct} (Is fully correct with, possibly, extra information). We create a special test set of $92$ challenging queries: $20$ factual queries adapted from SimpleQA and CWQ, $17$ verbose factual queries developed for this research, $15$ recent factual queries answerable with $\sim$ one-year-old data, $20$ 'yes/no' reasoning queries, and $20$ factual reasoning queries from CWQ.

The annotation results for all methods are presented in Table \ref{tab:human_eval}. \ours w/KG + Web reports a significant improvement of 21\% over the baseline WebGLM. ToG has competitive performance as WebGLM but has a higher proportion of "IDK" answers. We believe this is due to the reliance on ChatGPT's parametric knowledge when beam search fails.

\begin{table}[h]
\centering
\resizebox{\linewidth}{!}
{\begin{tabular}{lccc}
\hline
\textbf{Model} & \textbf{IDK} & \textbf{Incorrect} & \textbf{Correct}\\ \hline
\texttt{IO Prompt w/ChatGPT} & $0.38$ & $0.21$ & $0.41$ \\
\texttt{CoT w/ChatGPT} & $0.41$ & $0.18$ & $0.40$ \\ \hline
\texttt{ToG} & $0.25$ & $0.25$ & $0.50$ \\
\texttt{WebGLM} & $0.00$ & $0.47$ & $0.53$ \\ 
\texttt{\ours w/KG} & $0.00$ & $0.48$ & $0.52$ \\
\texttt{\ours w/Web} & $0.01$ & $0.41$ & $0.58$ \\
\texttt{\ours w/KG + Web} & $0.00$ & $0.26$ & $\mathbf{0.74}$ \\
\hline
\end{tabular}}
\caption{Human evaluation performance measured in Correctness (\textit{is it able to correctly answer the question?}) on a small 92 hand-picked challenging queries dataset using 3 annotation labels: \texttt{IDK} (unable to answer), \texttt{Incorrect}, \texttt{Correct}.}
\label{tab:human_eval}
\end{table}

\begin{table*}[h]
\centering

\small
\begin{tabular}{c|c|ccc|ccc}
\hline
 & & & \textbf{ELI5} & & & \textbf{NQ} & \\
\textbf{Quotes} & \textbf{Retriever} & \textbf{Per} & \textbf{AS} & \textbf{SC} & \textbf{Per} & \textbf{AS} & \textbf{SC} \\
\hline
\textbf{Top-1} & \texttt{WebGLM} & $1.76$ & $0.37$  & $0.35$  & $2.07$  & $0.28$  & $0.49$   \\
& \texttt{ {EWEK-QA (EE)}} & $2.07$ & $0.48$ & $0.49$ & $1.76$  & $0.37$ & $0.35$ \\
& \texttt{ {EWEK-QA (PS)}} & $2.2$ & $0.57$ & $0.54$ & $2.51$  & $0.47$ & $0.73$ \\
& \texttt{ EWEK-QA (EE + PS)} & $2.23$ & $0.5$  & $0.58$  & $2.66$  & $0.53$  & $0.8$   \\
\hline
\textbf{Top-5} & \texttt{WebGLM } & $1.71$ & $0.35$ & $0.34$ & $1.86$ & $0.3$ & $0.38$ \\
& \texttt{ {EWEK-QA (EE)}} & $1.9$ & $0.45$ & $0.37$ & $2.2$  & $0.45$ & $0.57$ \\
& \texttt{ {EWEK-QA (PS)}} & $1.99$ & $0.49$ & $0.41$ & $2.17$  & $0.39$ & $0.55$ \\
& \texttt{ EWEK-QA (EE + PS)} & $2.02$ & $0.5$ & $0.43$ & $2.36$  & $0.47$ & $0.64$ \\
\hline
\end{tabular}
\caption{Human evaluation of the web quotes retrieved by our system and WebGLM measured using Pertinence (Per), Answer Span (AS), and Self-containment (SC). Evaluated by professional human annotators on 200 random queries ELI5 and Natural Questions datasets.}
\label{tab:quote_scores}
\end{table*}

\subsection{Web Retriever Evaluation} \label{sec:web-retriever-human-eval}
To measure the quality of \ourret  quotes, we obtain human annotations for the top quotes retrieved by \ourret and the existing WebGLM retriever on 100 random queries from two datasets: ELI5 \citep{fan2019eli5} and Natural Questions \citep{kwiatkowski2019natural}. The results are presented in Table~\ref{tab:quote_scores}. The quotes are annotated along three dimensions: {\it Answer Span} (AS), {\it Self-Containment} (SC), and {\it Pertinence} (Per). AS measures what proportion of the quote is used verbatim by the LLM Answer Composer, SC measures whether the answer in the quote is complete or truncated, and Pertinence measures how relevant the quote is to the query and whether it is able to answer the query (annotated on a scale of $0-3$). Refer to \S\ref{sec:quote-human-eval-appendix} for more details.

As can be seen in Table~\ref{tab:quote_scores}, the quality of our quotes is significantly better than WebGLM quotes on both datasets across all three dimensions. This holds true whether we consider only the topmost quote or all top five quotes extracted by the retriever. This can be attributed to our improved heuristic parser and the additional evidence extractor component that is able to extract more relevant and complete quotes from webpages.

\subsection{Ablation and Analysis}

\subsubsection{Efficiency Analysis.}
\label{sec:eff_analysis}

In Table \ref{tab:efficiency}, we study the efficiency of \ours over ToG. In order to ensure a fair comparison, we stick to open-source LLMs for both methods. We observe that ToG's performance significantly degrades when using small open-source LLMs (i.e. {LLaMA-2}-13B). On the other hand, \ours uses even smaller open sourced LLMs and achieves SoTA performance. Moreover, \ours requires only one LLM call to compose an answer as opposed to ToG which can require up to $2ND + D + 1$, where $N$ and $D$ are the width and depth of the beam search respectively. Our method relies on fast retrievers and small embedding models (e.g. SentenceBert) allowing for $5\times$ decrease in LLM calls per question.

\begin{table*}[h]
\centering
\resizebox{0.8\textwidth}{!}{\begin{tabular}{c|ccccc}
\hline
\textbf{Dataset}                       & \textbf{Method}     & \textbf{LLM}        & \textbf{Avg. Runtime (s)} & \textbf{Avg. \# LLM calls} & \textbf{Hits@1} \\ \hline
\multirow{3}{*}{WebQSP}       & \texttt{ToG}                                      & {LLaMA-2}-13B  & $128.7$              & $5.6$               & $45.6$                \\ \cline{2-6} 
                              & \texttt{WebGLM}                                      & WebGLM-10B  & $44$            & $1$               & $65$                \\ \cline{2-6} 
                              & \multirow{3}{*}{\texttt{\ours} (Ours)} & {LLaMA-2}-13B & $29$               & $1$                 & $73.2$                \\
                              &                                         & WebGLM-10B & $40$               & $1$                 & $72.9$                \\
                              &                                          & WebGLM-2B  & $21$               & $1$                 & $68.9$                \\ \hline
\multirow{3}{*}{WebQuestions} & \texttt{ToG}                                      & {LLaMA-2}-13B  & $124.4$            & $5.7$               & $37.8$                \\ \cline{2-6} 
                              & \texttt{WebGLM}                                      & WebGLM-10B  & $45$            & $1$               & $54.3$                \\ \cline{2-6} 
                              & \multirow{3}{*}{\texttt{\ours} (Ours)} & {LLaMA-2}-13B & $26$               & $1$                 & $60.8$                \\
                              &                                          & WebGLM-10B & $35$               & $1$                 & $61.2$                \\
                              &                                          & WebGLM-2B  & $20$               & $1$                 & $58.4$                \\ \hline
\end{tabular}}
\caption{Efficiency vs Performance Analysis: Comparing the baseline ToG with our method locally using the {LLaMA-2} and WebGLM models respectively. \textit{"Avg. \# LLM calls"} represents the average number of LLM calls performed by ToG during search and reasoning stages per question. \textit{"LLM"} represents the LLM used by the method to predict the answer. Here, \ours uses both KG and web external knowledge. \textit{"Hits@1"} reports the performance on 1000 samples.}
\label{tab:efficiency}
\end{table*}

\begin{table*}[h]

{
\resizebox{\linewidth}{!}{

\begin{tabular}{lrrrrrr} 
& Total Time & Extract (Split) & Fetch (Crawling) & Filter (rank) & Search Engine \\
\hline WebGLM Retriever & 12.2093 & 0.785376 & 8.44552 & 1.8486 & 1.12285 \\
& & & & & \\
& Total Time & Deduplicator & EE + PS & Retrieve \& Re-rank & Search Engine \\
\hline EWEK-QA Web Retriever & 11.4965 & 0.12887 & 9.47622 & 1.0043 & 1.0028
\end{tabular}
}}

\caption{{Average run time (in seconds) of each module within EWEK-QA web retriever and WebGLM retriever for a single query. EE and PS refer to Evidence Extractor and Paragraph Splitter, respectively. The experiment was performed on 230 questions randomly selected from the Natural Questions dataset.}}
\label{tab:time_analysis_rets}
\end{table*}

In Table \ref{tab:efficiency}, we also present an analysis of how variations in the answer composer model impact the performance of our system. This examination leverages two distinct datasets: WebQSP, a KGQA dataset, and WebQuestions, an ODQA dataset. Specifically, we conducted experiments employing three different models: WebGLM-10B, WebGLM-2B, and {LLaMA-2}-13B. The {LLaMA-2}-13B (chat version) model is fine-tuned on the WebGLM-QA dataset \citep{webglm} \footnote{https://huggingface.co/datasets/THUDM/webglm-qa}. This dataset, previously utilized by \citet{webglm} for fine-tuning WebGLM-2B/10B answer composer models, comprises of questions, reference passages, and corresponding answers grounded within these passages. Fine-tuning of the {LLaMA-2}-13B model was conducted across 8 V100 GPUs. More details can be found in  \S\ref{app:answer_composer}.

Our findings indicate that the performance of our system remains robust across various off-the-shelf open-source LLMs serving as answer composers. Notably, despite similarities in LLM utilization between our approach and the baselines, our method enhances the quality of the extracted knowledge and outperforms both in terms of effectiveness and runtime.

{Table \ref{tab:time_analysis_rets} provides a comprehensive runtime (sec/query) comparison between the EWEK-QA web retriever and WebGLM retriever, conducted on a subset of 230 questions from the Natural Question dataset. We have run WebGLM with the default hyperparameters provided by the authors. We ran both EWEK-QA and WebGLM retriever models on V100 Nvidia GPUs.
The efficiency of our web retriever stems from our deliberate design choices. Our design facilitates parallelism during web page scraping and segmentation through the Evidence Extractor and Paragraph Splitter modules, as these modules operate independently. Also, each web page is parsed and segmented independently of others, optimizing efficiency. 
Unlike employing a single retriever or re-ranker, we adopt a double-module format for faster inference. In this format, a smaller, faster language model (LM) with 22 million parameters filters out noisy quotes, retaining only the most promising 70 quotes, which are then reranked by a larger LM with 900 million parameters to enhance final ranking. Notably, this reranking process with the larger model is swift due to the small number of quotes involved in a single forward pass. }

\begin{figure}[h]
    \centering
    \includegraphics[width=0.9\linewidth,clip,trim = 0.5cm 0cm 1cm 1cm]{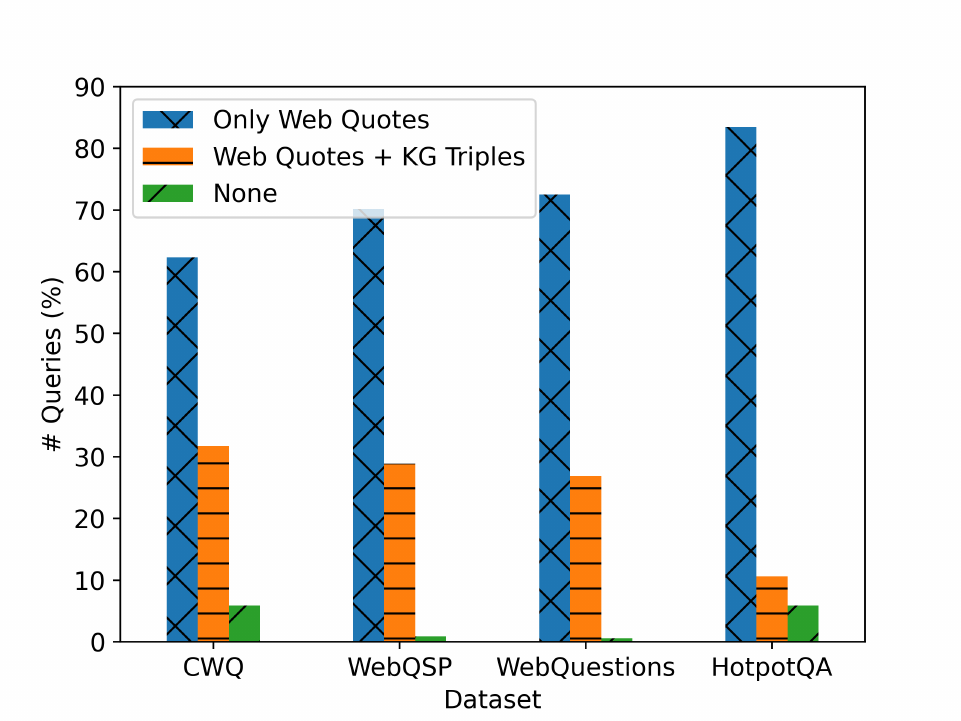}
    \caption{Sources of the quotes cited by the Answer Composer across queries from two KGQA and two ODQA datasets. "None" denotes that the answer contains no citations. "Web Quotes + KG Triples" indicate that both KG Triples and Web Quotes are cited.}

    \label{fig:citation_analysis}
\end{figure}

\subsubsection{Citation Analysis}

In order to study the efficacy and usefulness of our knowledge extraction approach, we conduct an analysis on the quotes cited by the LLM Answer Composer for queries from two KGQA and two ODQA datasets. Figure \ref{fig:citation_analysis} presents the distribution of citation sources in the \ours answers across the four datasets. Interestingly, the model relies only on the Web Quotes for most of the queries across all datasets (70\% of WebQSP questions). 
We attribute this to the richness and diversity of information available on the Web as compared to knowledge graphs. 
Nevertheless, KG Triples are utilized by the model for a large number of queries (29\% of WebQSP questions).

\paragraph{Citation Accuracy.} As hallucination is a common problem with LLMs, we verify the accuracy of these citations. We extract cited sentences from the answer and use GPT-3.5 to assess whether it is supported by the corresponding citation. We achieve an average citation accuracy of $89.6$\% across the four datasets, thereby strengthening the claim that hallucination is not biasing the citation analysis. Refer to \S\ref{sec:citation-appendix} for per-dataset accuracy, prompt details and an analysis on the number of cited quotes.

\section{Conclusion}

We introduce EWEK-QA, an efficient and generic QA system that is capable of answering both open domain and multi-hop reasoning questions. Contrary to prior works, our system relies on two external knowledge modalities: KGs and the web. We develop an adaptive web retriever to extract coherent and complete quotes from webpages. Furthermore, our ToG-E method eschews reliance on LLMs to extract the most relevant KG triples. Extensive experimental results on a variety of dataset types show significant efficiency gains over baselines. For future work, we aim to further improve the KG subgraph extraction module through more powerful embedding methods and experiment with using bigger backbone LLM models.

\section*{Limitations}

The datasets used in this work successfully benchmark the multi-hop reasoning abilities of all methods. However, we have found the exact match (i.e. Hits@1) evaluation to be constraining for this setting. For example, there exists many cases where the answer composer will correctly answer a question but in a different wording than expected. Therefore, it is worthwhile to scale up the human evaluation to gain more reliable results. Moreover, we do not utilize the most up-to-date KGs such as WikiData\footnote{https://www.wikidata.org/wiki/Wikidata:Introduction} which can limit our performance on temporal questions.

\bibliography{anthology,custom}
\bibliographystyle{acl_natbib}

\appendix

\section{Appendix}
\label{sec:appendix}

\subsection{Implementation Details}\label{app:imp_details}

Since initial topic entities for HotPotQA and NQ are not provided by default, we use the ReFinED model~\cite{ayoola-etal-2022-refined} to identify the WikiData~\cite{wikidata} entities in each question and map them to the corresponding FreeBase entities via the ``FreeBase ID" relation. We discard the questions where no Freebase entities are found. Moreover, we do not use the provided context for HotpotQA questions to test the retrieval performance of the baselines.

All method outputs were reproduced, except for DeCAF and KD-CoT. DeCAF outputs were obtained from their GitHub repository\footnote{https://github.com/awslabs/decode-answer-logical-form}, evaluated using our script, while KD-CoT numbers were extracted from their paper. All results are on $1000$ random samples except for WebQSP and CWQ which are on the full test set and for Natural Questions which is on random 400 samples.

We use {LLaMA-2} with ToG. {LLaMA-2}  was run on $8$ V$100$-$32$G GPUs without quantization, with temperature parameter $0.4$ for pruning and $0$ for the reasoning process. The maximum token length for the generation is set to $256$. We use $5$ shots in the ToG prompts for all the datasets. The maximum depth and width of the beam search is fixed to $3$. ToG-E uses the same hyper-parameters except that there is no reasoning/stopping stage and the pruning is performed via SentenceBert \cite{reimers-gurevych-2019-sentence}. The final answer is generated by the answer composer given the extracted KG triples and web quotes. For \ours and WebGLM, we retrieve the top $5$ relevant quotes of max length $128$ tokens. The web-pages are retrieved via the Bing search API.

\subsection{Computing Infrastructure}
All experiments were done on a Ubuntu 20.4 server with 72 Intel(R) Xeon(R) Gold 6140 CPU @ 2.30GHz
cores and a RAM of size 755 Gb. We use a NVIDIA Tesla V100-PCIE-32GB GPU.

\subsection{Citation Accuracy}
\label{sec:citation-appendix}
The citation accuracy on the individual datasets is presented in Table \ref{tab:citation-accuracy}. We used the following prompt with GPT-3.5 to assess if the citation was accurate or not:

{\small
\begin{spverbatim}
You are given an Answer and a Context. Your task is to identify whether the information in the Answer is present in (or supported by) the information in the Context. Output "Yes" if the Answer is supported by the Context.
    Answer: {sub-answer}
    Context: {quote}
\end{spverbatim}}

\begin{table}[]
\centering
\begin{tabular}{|c|c|}
\hline
\textbf{Dataset}          & \textbf{Accuracy} \\ \hline
CWQ          & 86.6 \\
WebQSP       & 92.7 \\
WebQuestions & 93.2 \\
HotpotQA     & 86.1 \\ \hline
\end{tabular}
\caption{Citation Accuracy for the four datasets. GPT-3.5 is used for judgements.}
\label{tab:citation-accuracy}
\end{table}

\begin{figure}[h]
    \centering
    \includegraphics[width=\linewidth,clip,trim = 0.5cm 0cm 1cm 1cm]{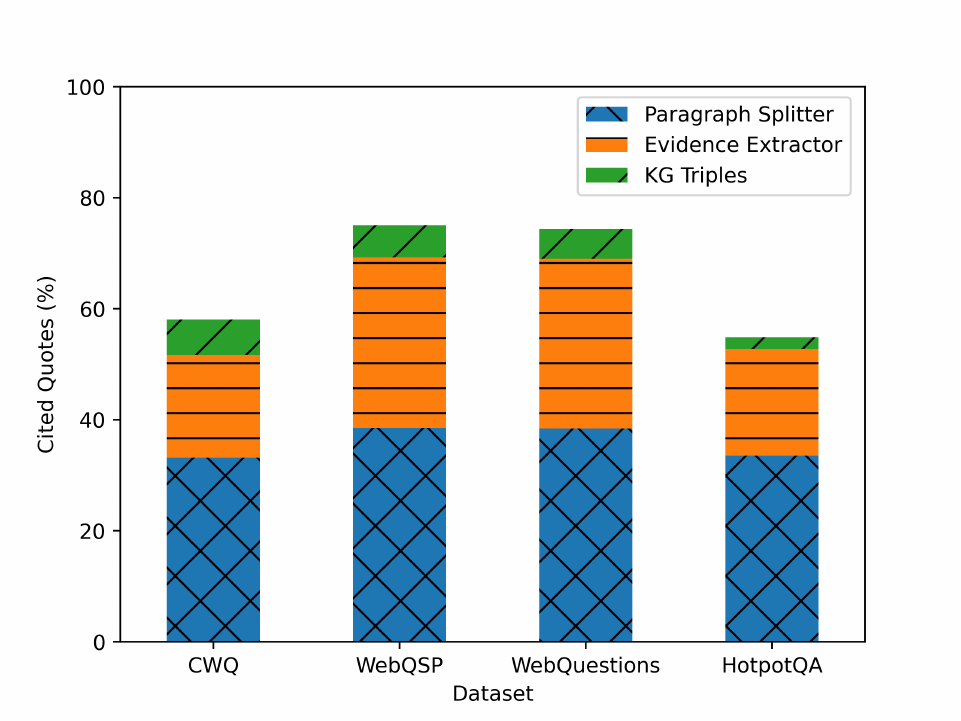}
        \caption{This figure depicts what percentage of the quotes provided by our knowledge extraction approach are cited in the final answer. Note that Paragraph Splitter and Evidence Extractor correspond to Web Quotes, and KG Triples come from the knowledge graph. }
    \label{fig:cited_quotes}
\end{figure}

We also analyse \textit{how many} of the quotes provided by our knowledge extraction approach are deemed useful by the answer composer. Figure \ref{fig:cited_quotes} shows that, on average, 65\% of the quotes provided are cited by the answer composer across the four datasets. More importantly, quotes originating from both streams of our adaptive web retrieval -- Paragraph Splitter and Evidence Extractor -- are used in the final answer.

\subsection{Answer Composer}\label{app:answer_composer}
Given the quotes, we prompt the WebGLM $10$B answer composer model with the following:
{\small
\begin{spverbatim}
[CLS] Reference [1]: {Quote1} \Reference [2]: {Quote2} \Reference [3]: {Quote3} \Reference [4]: {Quote4} \Reference [5]: {Quote5} \Question: {Question} \Answer: [gMASK] <|endoftext|> <|startofpiece|>
\end{spverbatim}}

When using the {LLaMA-2}-$13$B answer composer, we use the following prompt:

{\small
\begin{spverbatim}
<s> [INST] <<SYS>> Given the following quotes answer the question. You are given five quotes with their numbers. Each quote used in the answer should be cited with [ and ] symbols and the number of the quote in between.  <</SYS>><s> [INST] <|QUESTION|> {Question} <s> [INST] <|QUESTION|> <|QUOTES|> 
1: {Quote}
2: {Quote2}
3: {Quote3}
4: {Quote4}
5: {Quote5}
<s> <|ANSWER|>
\end{spverbatim}}

When KG triples are included, they are passed in as the first quote.

\subsection{Web Retriever Quotes Annotation}
\label{sec:quote-human-eval-appendix}
For this task, the annotators were asked to evaluate the quality of the extracted quotes. They were given files containing queries and quotes. To evaluate how good is the Quote at answering the Query or at contributing in answer the query, the annotators were asked to work on three different metric for each quote:
\begin{itemize}
    \item \textbf{Pertinence:} a score that measures how relevant the quote is to the query and whether the answer can be found in the quote. The given score must be $[0,3]\uparrow$:
    \begin{itemize}
        \item $0$ means that the quote does not answer the query AND the query-quote pair is irrelevant (different subjects).
        \item $1$ means that the quote does not answer the query BUT the query-quote pair is relevant (have same subject).
        \item $2$ means that the quote partially answers the query AND the query-quote pair is relevant (have same subject).
        \item $3$ means the quote completely answers the query AND the query-quote pair is relevant (have same subject).
    \end{itemize}
    \item \textbf{Answer-span:} this metric is used to know where the answer is. The annotators were asked to highlight the part of the quote which answer (even partly) the query. The highlighted text must be a single continuous string of text; if the answer to the query appears in multiple sections, separated by non-related data, all the sections must be highlighted. The score for this metric is computed by dividing the length (number of characters) of the highlighted text by the length (number of characters of the quote), so the score would be $[0,1]\uparrow$.
    \item \textbf{Self-containment:} this metric is used to measure if the answer was cut-off or absent. A binary score was assigned to each quote; $0$ means that the quote does not contain the answer or only partially, $1$ means that the quote correctly contains and mentions the answer.
\end{itemize}

See Tables \ref{tab:quote_eval_examples_eli5} and \ref{tab:quote_eval_examples_nq} for examples of human quote evaluation on queries from the Eli5 and NQ datasets respectively.

\subsection{Generated answer Annotation}
\label{sec:answer-human-eval-appendix}
For this task, the annotators were asked to evaluate the evaluate if and how correctly does the generated answer respond to the query. They were given files containing queries and generated answers. To do so, the annotators were asked to use a single metric:
\begin{itemize}
    \item \textbf{Correctness:} this metric indicates how well and how completely did the generated answer correctly respond the query? The given score must be $[0,2]\uparrow$: 
    \begin{itemize}
        \item $0$ means the generated answer indicates it does not know or is unable to answer the query.
        \item $1$ means the generated answer responds fully or partially incorrectly. This includes the answers that are only partially correct.
        \item $2$ means the generated answer responds correctly and might even include additional information.
    \end{itemize}
\end{itemize}

\subsection{Adaptive Web Retrieval} \label{sec:adaptive-web-appendix}
In this section, we layout additional details about the two components that produce candidate quotes, namely Paragraph Splitter and Evidence Extractor.
\subsubsection{Paragraph Splitter}
For every webpage returned by the search engine, we scrape the contents of each page using BeautifulSoup\footnote{https://www.crummy.com/software/BeautifulSoup/}. Like in \citet{webglm}, we divide the webpage contents into a list of candidate passages using line breaks. Since web-scraping is a time consuming task, we use multi-threading to scrape and parse each webpage in parallel. Additionally, for efficiency, we cache the search engine results and the scraped URL contents in a database.
We apply additional constraints to improve the quality of the candidate passages produced by the heuristic parser. In addition to using the newline character, we also utilise the \texttt{<p>} tags in the webpage's HTML to produce candidate passages; any passage that has less than $10$ tokens is discarded. If a passage contains more than $80$ tokens, we further break it down into shorter passages no longer than $80$ tokens such that the sentence boundaries are respected. 
This is slightly different from WebGLM's parser since they just relied on line breaks and used \texttt{html2text} to extract the text from HTML pages. Furthermore, in WebGLM's parser, lines shorter than $50$ characters were dropped and longer lines are truncated with first $1200$ characters followed by ``$\ldots$''.

\begin{figure}[h]
    \centering
    \includegraphics[width=1.0\linewidth]{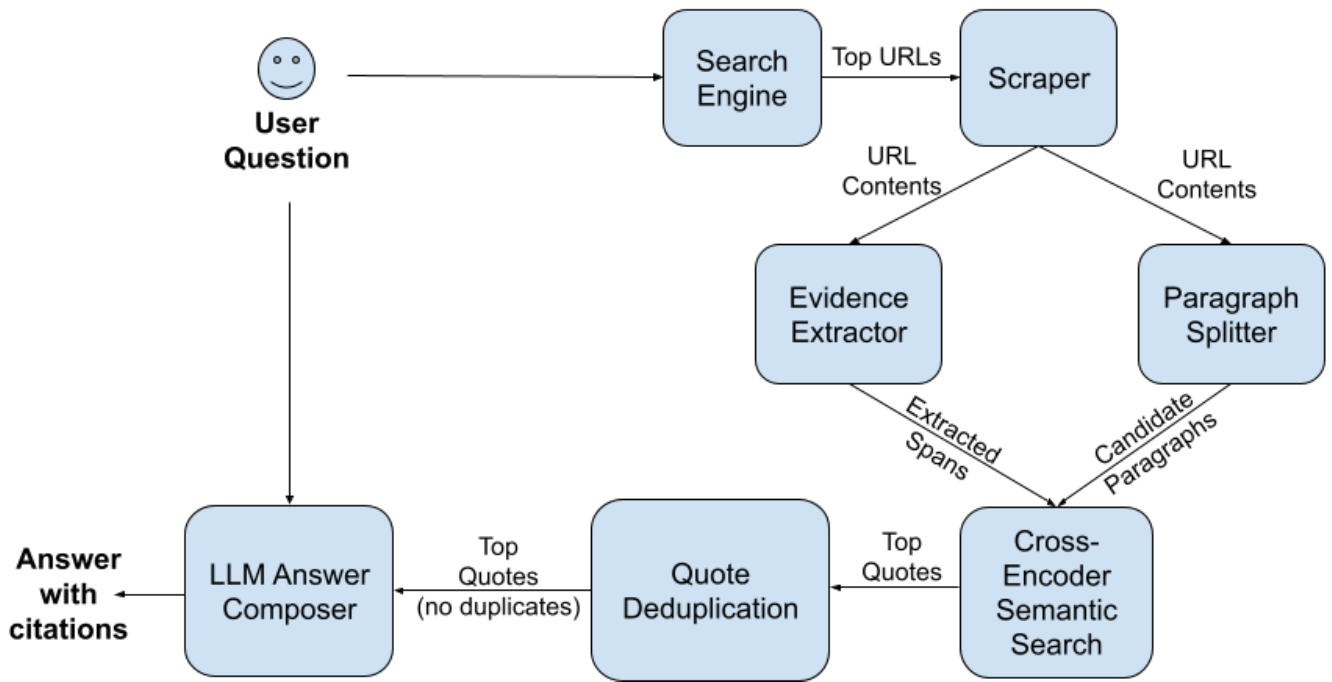}
        \caption{The pipeline for our adaptive web retrieval module.}
    \label{fig:adaptive_web_retriever}
\end{figure}

\begin{figure}[h]
    \centering
    \includegraphics[width=1.0\linewidth]{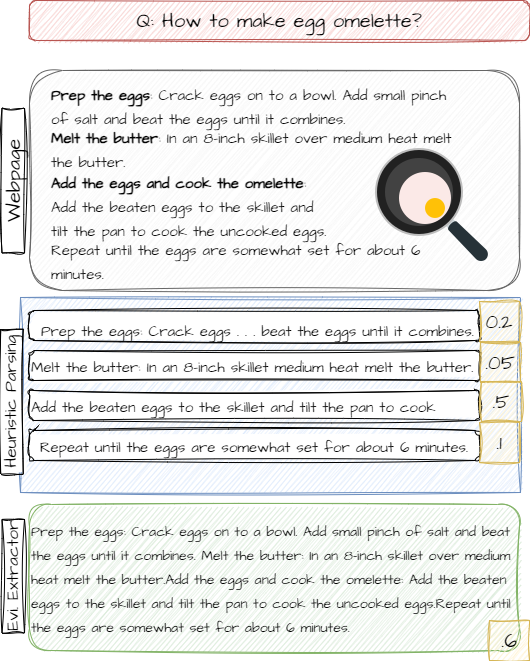}
        \caption{The candidate quotes produced by the Paragraph Splitter (Heuristic Parser) can be incomplete if they break on newlines or \texttt{<p>} tags. In comparison, a trained Evidence Extractor model can extract self-contained quotes.}
    \label{fig:evidence_heuristic}
\end{figure}

\subsubsection{Evidence Extractor}
\label{sec:apendix_EE}
By exploiting the similarity between evidence extraction and machine reading comprehension (MRC), we fine-tune the pre-trained MRC model--- DeBERTa \cite{he2021deberta}--- to extract a span of text from the documents. Instead of pursuing an answer span (like in MRC), the target here is to extract \textit{evidence spans} that can provide support to answer the question. In comparison to the Paragraph Splitter, the candidate passages produced by Evidence Extractor are likely to be more complete and \textit{self-contained} since it is not restricted to any predefined heuristic for chunking which can lead to incomplete chunks. See Figure \ref{fig:evidence_heuristic} for an example.
\paragraph{Training Data.} For this task, we use the MS Marco dataset which is a collection of about 1M queries sampled from Bing's search logs. The human editors are shown web passages relevant to the query and asked to compose a well-formed answer. Each query is accompanied with a set of 10 passages which may contain the answer to the question. The editors annotate the passage as \textit{is\_selected} if they use it to compose their final answer. Specifically, we create a training set of $110K$ instances from the {\it Train} split of MS Marco dataset using only the passages that the annotators tagged as useful while composing their answer (marked in the metadata with \textit{is\_selected} = $1$). Similar to span-prediction based MRC, each train instance is a three-tuple of $(q_i, s_i, c_i)$ where $q_i$ refers to the query, $s_i$ refers to the relevant passage span that the annotators used when composing their answer, and $c_i$ is the full text of the webpage ($u_i$) from where $s_i$ was derived. For every useful passage (\textit{is\_selected} = $1$) per query, we use it and the corresponding webpage with the question to create a data-point in our dataset. The model trained is expected to predict the span, $s_i$, when jointly provided with query and context webpage text ($q_i$,$c_i$).

\paragraph{Training Details.}
We experimented with DeBERTa and Longformer models for this task. We train the model for one epoch on four V100 GPUs using a learning rate of $3\times10^{-5}$ and training batch size of $32$. The best model is selected using the $f1$ measure on a set-aside validation set of $10k$ instances.

\subsection{Example Outputs}

A sample answer for \ours and several baselines can be found in Table \ref{tab:example_answers}. Table~\ref{tab:example_quotes} includes the KG triples and quotes extracted by \ours for a sample question in. Tables \ref{tab:quote_eval_examples_eli5} and \ref{tab:quote_eval_examples_nq} include examples of human quote evaluation on queries from Eli5 and NQ datasets respectively.

\begin{table*}[]
\centering
\resizebox{\textwidth}{!}
{\begin{tabular}{cp{9cm}}
\hline
\textbf{Model} & \textbf{Answer} \\

\hline
\multicolumn{2}{c}{\textbf{\textit{Which movie with a character called Ajila was directed by Angelina Jolie?}}} \\
\hline  
WebGLM & The movie with a character called Ajila was directed by Angelina Jolie and is called Girl, Interrupted. Angelina Jolie stars as Lisa Rowe, a sociopath in the 1960s psychiatric hospital who becomes Susanna Kaysen's unlikely friend on her journey to self-discovery[4].  \\
ToG & Based on the given knowledge triples, the movie directed by Angelina Jolie that features a character called Ajila is "In the Land of Blood and Honey". Therefore, the answer to the question is \{In the Land of Blood and Honey\}. \\
\ours & The movie directed by Angelina Jolie with a character called Ajila is In the Land of Blood and Honey[1][4]  \\

\hline
\end{tabular}}
\caption{Output example for the 3 models. The question is picked from the HotpotQA dataset.}

\label{tab:example_answers}
\end{table*}

\begin{table*}[]
\centering
\resizebox{.48\paperheight}{!}
{\begin{tabular}{cp{9cm}}
\hline
\textbf{Quote} & \textbf{Content} \\
\hline
\multicolumn{2}{c}{\textbf{\textit{Are the Laleli Mosque and Esma Sultan Mansion located in the same neighborhood?}}} \\
\hline  
KG Triples & {\small\texttt{('Esma Sultan Mansion', 'architecture.architect.structures\_designed', 'Balyan family'), ('Esma Sultan Mansion', 'architecture.structure.architect', 'Balyan family'), ('Laleli Mosque', 'religion.place\_of\_worship.religion', 'Islam'),('Balyan family', 'architecture.structure.architect', 'Esma Sultan Mansion'), ('Balyan family','architecture.structure.architect', 'Beylerbeyi Palace'), ('Balyan family', 'architecture.structure.architect', 'Dolmabahçe Mosque'),('Beylerbeyi Palace', 'architecture.architectural\_style.examples', 'Ottoman architecture'), ('Dolmabahçe Clock Tower', 'architecture.architectural\_style.examples', 'Ottoman architecture'), ('Dolmabahçe Clock Tower','architecture.structure.architectural\_style', 'Ottoman architecture')",'The Esma Sultan Mansion (Turkish: Esma Sultan Yals)}} \\
Quote 1 & The Esma Sultan Mansion (Turkish: Esma Sultan Yals), a historical yal located on the Bosphorus in the Ortakoy neighborhood of Istanbul, Turkey and named after its original owner Princess Esma Sultan, is used today as a cultural center after being redeveloped. \\
Quote 2 & The Laleli Mosque (Laleli Camii) is the centerpiece of the Laleli neighborhood in Istanbul , Turkey . It sits along Ordu Street (Ordu Caddesi), which is part of the historic Divan Yolu  \\
Quote 3 & The Laleli Mosque (Turkish: Laleli Camii, lit. 'Tulip Mosque') is an 18th-century Ottoman imperial mosque located in Laleli, Fatih, Istanbul, Turkey. \\
Quote 4 & Laleli is a neighborhood in Istanbul, Turkey, with a few points of interest. It's located in the Fatih district between Beyazt and Aksaray. Laleli Laleli runs along Ordu Street (Ordu Caddesi), which is part of the historic Divan Yolu. \\
Quote 5 & Location At the intersection of Ordu and Fethi Bey Streets in Laleli Neighborhood, Eminonu District., Istanbul, Turkey Directions Associated Names Mustafa III, Ottoman Sultan Turkey patron Mehmet Tahir Aga Turkey architect/planner Istanbul Turkey place Events / AH damaged in earthquake in 1766/1179 AH Show all 3 Style Periods Ottoman Variant Names Laleli Kulliyesi Alternate transliteration Laleli Kulliye Variant. About Home Sites Authorities Collections Search Laleli Kulliyesi Istanbul, Turkey The Laleli Complex is a mosque complex commissioned and partly designed by Ottoman sultan Mustafa III. The word "Lale", in Turkish, is a tulip. \\
\hline
\end{tabular}}
\caption{KG triples and quotes extracted by \ours. }
\label{tab:example_quotes}
\end{table*}

\begin{table*}[]
\centering

{\begin{tabular}{cp{9cm}ccc}
\hline
\textbf{Retriever} & \textbf{Quote} & \textbf{Per} & \textbf{AS} & \textbf{SC} \\
\hline
\multicolumn{5}{c}{\textbf{\textit{who was the king of england in 1756}}} \\
\hline  
WebGLM & King of Great Britain and Ireland from 1727 to 1760 & 1 & 0.0 & 0 \\
Ours & George II (George Augustus; German: Georg August; 30 October / 9 November 1683[a] - 25 October 1760) was King of Great Britain and Ireland, Duke of Brunswick-Luneburg (Hanover) and a prince-elector of the Holy Roman Empire from 11 June 1727 (O.S.) until his death in 1760. & 3 & 1.0 & 1 \\
\hline
\multicolumn{5}{c}{\textbf{\textit{what is the name of the hyena in lion king}}} \\
\hline  
WebGLM & In the first book of the novelization series, The Lion King: Six New Adventures, A Tale of Two Brothers, the young hyenas appear at the beginning where they surround a helpless Rafiki who had just arrived in the Pride Lands. King Ahadi comes to his rescue just in time, with his sons Mufasa and Taka (Scar) in tow, and demands that the hyenas leave. Later, Scar meets with the hyenas, the story revealing that he asked for their help. Shenzi advises Scar to make himself look good by making Mufasa look bad. Taka is intrigued by the suggestion and plans to incite a cape buffalo called Boma to fight Mufasa. His plan goes badly and his eye is wounded by the buffalo herd in the process, hence the nickname "Scar". & 1 & 0.0 & 0 \\
Ours & According to Timon in The Lion King 1, her full name is Shenzi Marie Predatora Veldetta Jacquelina Hyena. Her first name means \"savage\", \"pagan\", \"uncouth\", or \"barbarous\" in Swahili. Her distinguishing features are five prominent bangs hanging over her face and a mane that reaches all the way to her bangs. & 3 & 0.34 & 1 \\
\hline
\multicolumn{5}{c}{\textbf{\textit{who is the biggest selling female group of all time}}} \\
\hline  
WebGLM & Opening up the top three best-selling girl groups of all time is the American R\&B trio TLC. The girl group was formed in 1991 and it consisted of T-Bozm Chilli, Left Eye, and Crystal Jones, however, Crystal left the group early on. & 1 & 0.0 & 0 \\
Ours & The Spice Girls (here in 2008) are the best-selling girl group in history. & 3 & 1.0 & 1 \\
\hline
\end{tabular}}
\caption{Examples of human quote evaluation on queries from NQ.}
\label{tab:quote_eval_examples_nq}
\end{table*}

\begin{table*}[]
\centering

{\begin{tabular}{cp{9cm}ccc}
\hline
\textbf{Retriever} & \textbf{Quote} & \textbf{Per} & \textbf{AS} & \textbf{SC} \\
\hline  
\multicolumn{5}{c}{\textbf{\textit{Why do we prefer cold pillows to warm pillows?}}} \\
\hline  
WebGLM & - Firmness: Preferred firmness will differ from person to person. Firm pillows can provide greater support for the head and neck, while softer pillows can reduce pressure. A person can consider choosing pillows from brands that offer sleep trials or allow returns so they can replace or exchange their pillow if it is not comfortable. & 1 & 0.0 & 0 \\
Ours & Is it better to sleep on a cold pillow? Research shows that a warm environment may keep people awake and disrupt the body's regular sleep-wake cycle, resulting in poor quality sleep. A cold pillow may help a person feel cool enough to fall asleep. & 3 & 0.84 & 1 \\
\hline
\multicolumn{5}{c}{\textbf{\textit{PayPal, how it works, if it's trustworthy/worthwhile and should I get it?}}} \\
\hline
WebGLM & PayPal has various measures in place to make sure PayPal is safe for both buyers and sellers, whenever they make or receive an online payment. It’s one of the reasons PayPal grew to be such a popular payment platform. For buyers, PayPal offers protection on eligible PayPal purchases, as well as 24/7 monitoring of accounts to watch out for suspicious transactions, dispute resolution services and more. & 2 & 1.0 & 0 \\
Ours & PayPal is a service that allows you to use your credit card to pay for things on the internet. It's not trustworthy or worthwhile, but if you're willing to put in the time and effort to learn how to use it, it's worth it. If you don't trust PayPal, don't use it. & 3 & 1.0 & 1 \\
\hline
\end{tabular}}
\caption{Examples of human quote evaluation on queries from Eli5.}
\label{tab:quote_eval_examples_eli5}
\end{table*}

\end{document}